\documentclass[letterpaper]{article}
\usepackage{aaai}
\usepackage{cancel}
\usepackage{times}
\usepackage{helvet}
\usepackage{courier}
\usepackage{color}
\usepackage[utf8]{inputenc} 
\usepackage[T1]{fontenc}    
\usepackage{hyperref}       
\usepackage{url}            
\usepackage{booktabs}       
\usepackage{amsmath}
\usepackage{amsfonts}       
\usepackage{mathtools}
\usepackage{nicefrac}       
\usepackage{microtype}      
\usepackage{bm}
\usepackage{xfrac}
\usepackage{lipsum}
\usepackage{graphicx}
\usepackage{subfigure}
\usepackage{caption}
\usepackage{algorithmicx}
\usepackage[group-separator={,}]{siunitx}
\usepackage[ruled,vlined]{algorithm2e}
\usepackage{cleveref}
\usepackage{upgreek}
\usepackage{soul}
\usepackage{multirow}
\crefformat{section}{\S#2#1#3} 
\crefformat{subsection}{\S#2#1#3}
\crefformat{subsubsection}{\S#2#1#3}
\newcommand{\res}{\mathrm{r}}
\newcommand{\x}{\mathrm{x}}
\newcommand{\X}{\mathrm{X}}
\newcommand{\y}{\mathrm{y}}

\newcommand{\Y}{\mathrm{Y}}

\newcommand{\Proj}{\mathrm{P}}
\newcommand{\Q}{\mathrm{Q}}
\newcommand{\R}{\mathrm{R}}
\newcommand{\I}{\mathrm{I}}
\newcommand{\C}{\mathrm{C}}
\newcommand{\G}{\mathrm{G}}
\newcommand{\Cd}{\C^\dagger}

\newcommand{\Ct}{\widetilde{\C}}
\newcommand{\EE}{\mathbb{E}}
\newcommand{\RR}{\mathbb{R}}
\newcommand{\tr}{\text{tr}}
\newcommand{\nub}{\upnu}
\newcommand{\bet}{\upbeta}
\newcommand{\Th}{\Theta}
\newcommand{\Tht}{\widetilde{\Th}}
\newcommand{\Sig}{\Sigma}
\newcommand{\epst}{\tilde{\epsilon}}
\newcommand{\epstinv}{\epst^{\scriptscriptstyle -1}}
\newcommand{\etap}{\eta^\prime}

\newcommand{\mb}{\mathrm{m}}

\newcommand{\zero}{\mathbf{0}}

\newcommand{\mub}{\upmu}
\newcommand{\Sb}{\mathrm{\Lambda}}
\newcommand{\U}{\mathrm{U}}

\newcommand{\UP}{{\text{U}}}

\newcommand{\pgrad}{\overline{\nabla}}
\newcommand{\thet}{\uptheta}

\newcommand{\St}{\text{St}_{(d,k)}}

\def\argmin{\mathop{\mbox{argmin}}}

\newcommand{\Nc}{\mathcal{N}}

\newcommand{\Frob}{\text{\tiny F}}

\makeatletter
\newcommand*{\Rbar}{}%
\DeclareRobustCommand*{\Rbar}{%
  \mathpalette\@Rbar{}%
}
\newcommand*{\@Rbar}[2]{%
  \sbox0{$#1\mathrm{\R}\m@th$}%
  \sbox2{$#1\R\m@th$}%
  \rlap{%
    \hbox to\wd2{%
      \hfill
      $\overline{%
        \vrule width 0pt height\ht0 %
        \kern\wd0 %
      }$%
    }%
  }%
  \copy2 %
}
\makeatother

\newcommand{\citet}[1]{\citeauthor{#1} et al. \citeyear{#1}}

\begin{document}
%
\title{An Implicit Form of Krasulina's $k$-PCA\\ Update without the Orthonormality Constraint}
\author{%
 Ehsan Amid \quad Manfred K. Warmuth\\
  University of California, Santa Cruz\\
  Google Brain, Mountain View\\
  \texttt{\{eamid, manfred\}@google.com}
}
\maketitle
\begin{abstract}
\begin{quote}
  We shed new insights on the two commonly used updates for
    the online $k$-PCA problem, namely, Krasulina's and
    Oja's updates. We show that Krasulina's update
    corresponds to a projected gradient descent step on the
    Stiefel manifold of the orthonormal $k$-frames,
    while Oja's update amounts to a gradient descent step using the unprojected gradient.
    Following these observations, we derive a more
    \emph{implicit} form of Krasulina's $k$-PCA
    update, i.e. a version that uses the information of the
    future gradient as much as possible.  Most
    interestingly, our implicit Krasulina
    update avoids the costly QR-decomposition step
    by bypassing the orthonormality constraint. We show
    that the new update in fact corresponds to an online EM
    step applied to a probabilistic $k$-PCA model. The probabilistic view of the
    updates allows us to combine multiple models in a
    distributed setting. We show experimentally
    that the implicit Krasulina update yields
    superior convergence while being significantly faster.
    We also give strong evidence that the new update
    can benefit from parallelism and is more stable
    w.r.t. tuning of the learning rate.
\end{quote}
\end{abstract}

\section{Introduction}
Principal Component Analysis (PCA)~\cite{pearson} is one of
 the most widely used techniques in data
 analysis~\cite{elements}, dimensionality
 reduction~\cite{dr}, and machine learning~\cite{jolliffe}.
 The problem amounts to finding projections of the
 $d$-dimensional data along $k < d$ orthogonal directions
 such that the expected variance of the reconstruction
 error is minimized. Formally, let $\y \in \RR^d$ be a
 zero-mean random variable\footnote{We focus on the
 centered $k$-PCA problem since handling the mean value of a
 random variable is rather simple.}. In the vanilla $k$-PCA
 problem we seek a $d\times d$ projection
 matrix\footnote{Projection matrices are symmetric ($\Proj
 = \Proj^\top$) and idempotent ($\Proj^2 = \Proj$).}
 $\Proj$ of rank-$k$ such that the \emph{compression loss} of $\y$,
\begin{equation}
\begin{aligned}
\label{eq:pca-expectation-form}
    \ell_{\text{\tiny comp}}(\Proj)
    & =  \EE\big[\Vert \Proj\, \y-\y\Vert^2\big]= \tr\big((\I_d-\Proj)\,\EE[\y\,\y^\top]\big),\!\!\!\!
\end{aligned}
\end{equation}
is minimized. Here $\I_d$ denotes the $d\times d$ identity matrix.

 The optimum projection matrix can be
 decomposed\footnote{Note that this decomposition is not
 unique. However, there exists a unique decomposition in
  terms of the orthonormal eigenvectors of $\Proj^*$.} as $\Proj^* =
 \C\,(\C^\top\C)^{-1}\C^\top \coloneqq \C\,\Cd$, where $\C$
 is a $d\times k$ matrix that spans the space of top-$k$
 eigenvectors of the data covariance matrix
 $\EE\big[\y\,\y^\top\big]$. For a given
 $\C$, we denote by $\x \coloneqq \Cd\,\y
 \in \RR^k$, the projection of $\y$ onto the column
 space of $\C$. The reconstruction of $\y$ from the projection
 $\x$ can be calculated as $\C\,\x \in
 \RR^d$. In practice, \eqref{eq:pca-expectation-form} is
 approximated using a finite number of samples
 $\{\y_n\}_{n=1}^N$ resulting in the following objective:
\begin{equation}
\begin{aligned}
    \hat{\ell}_{\text{\tiny comp}}(\Proj) &=  \sfrac{1}{N}\,\tr\big((\I_d-\Proj)\,\sum_n \y_n \,\y_n^\top\big)\\
    &= \sfrac{1}{N}\,\tr\big((\I_d-\Proj)\,\Y\,\Y^\top\big)\,.
    \label{eq:pca-sample-form}
\end{aligned}
\end{equation}
Here, $\Y$ denotes the $d\times N$ matrix of observations,
i.e. the $n$-th column is $\y_n$
We also define $\X$ as the matrices of the
projected samples. Note that the
objective~\eqref{eq:pca-sample-form} is linear and thus
convex in $\Proj=\C\Cd$, but non-convex in $\C$. Nevertheless,
the solution can be efficiently calculated using the SVD of the empirical
data covariance matrix $\Y\Y^\top$.

Although the original PCA problem concerns the full-batch
setting, in many cases the dataset might be too large to be
processed by a batch solver. In such scenarios, online PCA
solvers that process a mini-batch (or a single observation)
at a time are more desirable.  In the online setting, two
elegant solutions proposed by Krasulina~\cite{krasulina}
and Oja~\cite{oja} for finding the top-$1$ direction have
received considerable attention and been studied
extensively throughout the years~\cite{chen,balsubramani,jain,allen}. In the following, we review the generalizations of these two algorithms to the $k$-PCA problem.

We show that both the Krasulina and Oja updates correspond to
stochastic gradient descent steps on the compression loss (and its reduced form)
using the gradient of the loss at the old parameter.
Using the terminology introduced in \cite{pnorm}, we call updates that are based on the old gradient the \emph{explicit} updates.
In this paper, we focus on deriving a more \emph{implicit} form of Krasulina's updates.
Implicit here means that the updates are based on the
future gradients at the updated parameters.
As an example, consider the following regularized loss minimization over the parameter $\thet$
\begin{equation}
\label{eq:gd}
    \thet_{t+1} = \argmin_{\thet} \big(\sfrac{1}{2\eta}\,
    \Vert\thet - \thet_t\Vert^2  + \text{loss}(\thet)\big)\, ,
\end{equation}
with \emph{learning rate} $\eta > 0$. Setting the derivative w.r.t. $\thet$ to zero yields the following gradient descent update
\begin{equation}
    \label{eq:implicit-gd}
    \thet_{t+1} = \thet_t - \eta\,  \nabla
    \text{loss}(\thet_{t+1}).
    \end{equation}
The above update is referred to as an implicit gradient descent update since it uses the gradient of the loss at the future parameter $\thet_{t+1}$. In cases where solving the update using the gradient at the future estimate is infeasible, the update is usually approximated by the gradient at the current estimate
\begin{equation}
    \label{eq:explicit-gd}
    \thet_{t+1} \approx\, \thet_t - \eta\,  \nabla \text{loss}(\thet_t)\, ,
\end{equation}
which is referred to as the explicit gradient descent
update. In many settings, implicit updates are more stable and have better convergence properties compared to their explicit counterparts~\cite{hassibi,pnorm}.

In summary:
\begin{itemize}
    \item We first formulate the Krasulina and Oja updates as an online (un)projected gradient descent steps on the Stiefel manifold.
    \item Using this observation, we derive a more implicit form for  Krasulina's update for the $k$-PCA problem that avoids the orthonormality constraint
	and is strikingly simple.
    \item We show that the new implicit Krasulina
	update actually amounts to an online EM step on
	a probabilistic $k$-PCA model. This allows combining
	multiple $k$-PCA models in a distributed setting using
	the recent framework of~\cite{online_em}.
    \item With an extensive set of experiments, we show
	that the implicit Krasulina update yields better convergence while being more stable w.r.t. the choice of initial learning rate. Furthermore, by avoiding the orthonormalization
	step and maintaining matrix pseudo-inverses instead, we achieve a much faster update.
	Further speedup can be achieved by running our algorithm in parallel
	and combining the results.
\end{itemize}
\subsection{Related Work}
Many efficient solvers have been developed for the vanilla PCA problem~\eqref{eq:pca-sample-form} throughout the years. For a reasonable data size, one can find the exact solution by applying a truncated SVD solver on the empirical data covariance matrix. Randomized SVD solvers~\cite{halko} and power methods~\cite{golub} are common alternatives when the size of the dataset is larger. Online PCA solvers are desirable when the data comes in as a stream or when a full pass over the data may not be feasible due to the large size of the dataset. In general,
the online algorithms iteratively perform the updates
using a mini-batch of observations (commonly a single observation) at every round.
Among the online algorithms, Oja's update~\cite{oja} and
its variants are the most well-studied~\cite{jain,allen}.
Noticeably, Shamir~\cite{shamir2015} showed exponential
convergence on a variance reduced variant of Oja's
algorithm. However, the algorithm requires multiple passes
over the data. On the other hand, a thorough convergence
analysis of Krasulina's algorithm and its extension to the
$k$-PCA problem is still lacking. A partial analysis was done recently in~\cite{tang} where
an exponential convergence was shown when the data
covariance matrix has low rank. However, this assumption might be too restrictive in real-world scenarios.
A lower expected rate of $\mathcal{O}(\sfrac{1}{t})$ still
holds for Krasulina's $1$-PCA update for high rank data~\cite{balsubramani}.
Other formulations for online PCA exploit the linearity of
the objective \eqref{eq:pca-expectation-form}
in $\Proj$ by essentially maintaining a mixture of
solutions for $\Proj$ of rank $k$ as a capped density matrix~\cite{warmuth2008,capped-msg}.
This approach leads to algorithms with optimal regret
bounds~\cite{nie} but these algorithms are fundamentally less efficient than the incremental counterparts that aim to optimize $\C$.
However, inspired by this approach of optimizing for $\Proj$,
an incremental heuristic method has been developed~\cite{incremental}
that gives reasonable experimental convergence, but again suffers
from high computational cost for the updates.

\section{Quadratic Program on the Stiefel Manifold}

Given the zero-mean random variable $\y \in \RR^d$,
consider the following optimization problem for the
centered $k$-PCA problem:
\begin{equation}
\begin{aligned}
\label{eq:residue-exp}
    \C^* &= \argmin_{\C\in\St}
    \ell_{\text{\tiny comp}}(\C)\,,\\
    \text{ for\,\,\, }  \ell_{\text{\tiny comp}}(\C)& \coloneqq
    \sfrac{1}{2}\,\EE\big[\Vert \C\C^\top\y-\y\Vert^2\big]\, ,
\end{aligned}
\end{equation}
where $\St = \{\C\,\in\,\RR^{d\times k}|\,\C^\top\C =
\I_k\}$ with $d \geq k$ is the compact Stiefel manifold of
orthonormal $d\times k$ matrices. We view $\St$ as an
embedded submanifold of $\RR^{d\times k}$. The
objective~\eqref{eq:residue-exp} is identical to the
expected compression loss~\eqref{eq:pca-expectation-form}.
Indeed when $\C\in\St$, then $\Proj = \C\C^\top$ is a
projection matrix. Also for $\C \in \St$, we have $\C^\top\C = \I_k$. Thus, we can rewrite~\eqref{eq:residue-exp} as
\begin{equation}
\begin{aligned}
\label{eq:var}
    \C^* &= \argmin_{\C\in\St}
    \ell_{\text{\tiny var}}(\C)\,,\\
    \text{ for\,\,\, }  \ell_{\text{\tiny var}}(\C) \coloneqq \sfrac{1}{2}\Big(\tr&\big(\EE[\y\y^\top]\big) - \tr(\C^\top\EE[\y\y^\top]\,\C)\Big)\, .
\end{aligned}
\end{equation}
Thus minimizing the compression
loss~\eqref{eq:residue-exp} is equivalent to maximizing the
variance $\tr(\C^\top\EE[\y\y^\top]\,\C) = \EE[\Vert\C^\top\y\Vert^2] = \text{Var}[\x]$
of the projection $\x = \C^\top\y$. Note that
although the values of the
objectives~\eqref{eq:residue-exp} and~\eqref{eq:var} are
identical when $\C \in \St$, they might yield different updates for the gradient based methods
as we shall see in the following.

A stochastic optimization procedure for solving~\eqref{eq:residue-exp} can be motivated similar to~\eqref{eq:gd} where an inertia term is added to the loss to keep the updated parameters close to the current estimates, that is,
\[
\C^{\text{\tiny new}} = \argmin_{\Ct\in\St}
\sfrac{1}{2}\,\Big(\sfrac{1}{\eta}\,\big\Vert \Ct -
\C\big\Vert^2_\Frob + \EE\big[\Vert\Ct\,\x-\y\Vert^2\big]\Big)\, ,\label{eq:qp-gd}
\]
where $\x = \Ct^\top\y$ is the projection onto the column space of $\Ct$ and $\Vert\mathrm{A}\Vert^2_\Frob = \tr(\mathrm{A}^\top\mathrm{A})$ denotes the squared Frobenius norm. A procedure for solving the optimization problem is based on iteratively applying a gradient descent step using the projected gradient of the objective onto the tangent space of $\St$ at $\C$ followed by a retraction step~\cite{liu-qp}. The tangent space at $\C\in \St$ is characterized by $T(\C) = \{\G \in \RR^{d\times k}\,|\, \G^\top\C + \C^\top\,\G = \zero_{k\times k}\}$ and the projected gradient of $ \ell_{\text{\tiny comp}}$, denoted by $\pgrad\,  \ell_{\text{\tiny comp}}$,
can be obtained by projecting the Euclidean gradient
$\nabla \ell_{\text{\tiny comp}}$ onto $T(\C)$:
 \begin{align*}
     \nabla \ell_{\text{\tiny
     comp}}&=\EE[(\C\C^\top\y-\y)\y^\top\C]\, ,\\
     \pgrad\, \ell_{\text{\tiny comp}}(\C) & = (\I_d - \C\,\C^\top)\, \nabla \ell_{\text{\tiny comp}}(\C)\\
 & = \C\,\C^\top\,\EE[\y\y^\top]\,\C-\EE[\y\y^\top]\,\C = \nabla \ell_{\text{\tiny
     comp}}\, .
 \end{align*}
Notice that $\pgrad\, \ell_{\text{\tiny comp}}(\C) = \nabla \ell_{\text{\tiny
     comp}}  \in
T(\C)$. In a stochastic approximation setting, the gradient
at each iteration is approximated by a given batch of
observations $\{\y_n\}_{n=1}^N$:
\begin{equation}
\begin{aligned}
    \pgrad\, \hat{\ell}_{\text{\tiny comp}}(\C) &=
    \sfrac{1}{N}\sum_n\big(\C\,\C^\top\,\y_n\y_n^\top\C - \y_n\y_n^\top\C\big)\\
    &= \sfrac{1}{N}\,(\C\X-\Y)\X^\top\, ,
\end{aligned}
\end{equation}
where $\Y\in \RR^{d\times N}$ denotes the matrix of observations and $\X\coloneqq \C^\top\Y$. Thus, the stochastic approximation update becomes
\begin{equation}
\begin{aligned}
\label{eq:krasulina-kpca}
    \Ct = \C - \eta\,  \pgrad\, \hat{\ell}_{\text{\tiny comp}}(\C)
    & = \C - \sfrac{\eta}{N}\, (\C\,\X-\Y)\,\X^\top\\
    \text{ and }\,\,\, \C^{\text{\tiny new}} & = \text{\texttt{QR}}(\Ct)\, .
    \end{aligned}
\end{equation}
The intermediate parameter $\Ct \notin \St$ in general,
but the second $\texttt{QR}$ step ensures that $\C^{\text{\tiny new}} \in \St$.
This update~\eqref{eq:krasulina-kpca} is identical to the extension of Krasulina's update to the $k$-PCA problem which was proposed recently in \cite{tang}.

Alternatively, the Euclidean gradient of~\eqref{eq:var} becomes
 \[
 \nabla \ell_{\text{\tiny var}}(\C) = -\EE\big[\y\,\y^\top\big]\C\,.
 \]
Projecting this gradient onto the tangent space yields
\[
\pgrad\, \ell_{\text{\tiny var}}(\C)
= (\I_d - \C\,\C^\top)\, \nabla \ell_{\text{\tiny var}}(\C) = \nabla \ell_{\text{\tiny comp}}(\C)\, .
\]
 Thus the update using the projected gradient
$\pgrad\, \ell_{\text{\tiny var}}(\C) $
 again yields Krasulina's update for $k$-PCA~\eqref{eq:krasulina-kpca}.
 Interestingly, the update using the Euclidean (unprojected) gradient
$\nabla \hat{\ell}_{\text{\tiny var}}(\C)$ gives
the well-known Oja update for $k$-PCA~\cite{oja}:
\begin{align*}
    \Ct = \C - \eta\, \nabla \hat{\ell}_{\text{\tiny var}}(\C)  & = \C + \sfrac{\eta}{N}\, \Y\,\Y^\top\,\C\, ,\\
    \text{ and }\,\,\, \C^{\text{\tiny new}} & = \text{\texttt{QR}}(\Ct)\, ,
\end{align*}
Thus, Oja's update is a
an stochastic approximation update that aims to maximize the variance of the projection, but ignores the structure of the tangent space of the Stiefel manifold.

Note that both the Krasulina and Oja updates are
explicit (similar to gradient descent update~\eqref{eq:explicit-gd}) in that they use the gradient at the old parameter $\C$ rather than the new parameter $\C^{\text{\tiny new}}$~\cite{pnorm}.
Unfortunately, using the new parameter $\C^{\text{\tiny
new}}$ for the updates as in~\eqref{eq:implicit-gd} does not yield a tractable solution.
However, we will show that a more implicit form of Krasulina's update,
i.e. the one that uses the gradient at the new parameter,
can be achieved when the orthonormality constraint is abandoned.

\section{New Update w.o. Orthonormality Constraint}

Instead of maintaining an orthonormal matrix $\C$,
we let $\C$ be any $d\times k$ matrix of rank-$k$ and use the fact that
the projection of an observation $\y\in\RR^d$ onto the column space of
$\C$ corresponds to  $\Cd\,\y\in\RR^k$:
\begin{equation}
    \C^{\text{\tiny new}}
    =
    \argmin_{\Ct}\,\sfrac{1}{2}\,\Big(\sfrac{1}{\eta}\,\big\Vert
    \Ct - \C\big\Vert^2_\Frob\\ + \EE\big[\Vert\Ct\,\x-\y\Vert^2\big]\Big)\, ,
    \label{eq:krasulina-obj}
    \end{equation}
    where $\x = \Cd\,\y$ is the projection using the old matrix $\C$.
    Setting the derivatives w.r.t. $\Ct$ to zero, we obtain
    \begin{align*}
	\C^{\text{\tiny new}} &= \C - \eta \EE\big[(\C^{\text{\tiny
    new}}\x -\y)\,\x^\top\big] \\
	&= \Big(\EE[\,\y \x^\top]  + \sfrac{1}{\eta}\, \C\Big)\Big(\EE[\,\x\x^\top] +  \sfrac{1}{\eta}\, \I_k\Big)^{-1} .
     \end{align*}
     For a batch of points $\{\y_n\}_{n=1}^N$,
     this new update, which we call the \emph{implicit Krasulina}
     update, becomes
     \begin{equation}
     \label{eq:k-pca-implicit}
     \boxed{\C^{\text{\tiny new}}  =\! \Big(\sfrac{1}{N}\, \Y\X^\top\!\! + \sfrac{1}{\eta}\, \C\Big)\Big(\sfrac{1}{N}\,\X\X^\top\!\! +  \sfrac{1}{\eta}\, \I_k\Big)^{-1}}\, .
    \end{equation}
    Note that an explicit variant can also be obtained by simply
    approximating $\C^{\text{\tiny new}}\,\X$ in the
    derivative equation by the old matrix $\C\,\X$:
    \[
    \C^{\text{\tiny new}} \approx \,\, \C \, - \, \,\sfrac{\eta}{N}\,\big(\C\,\X-\Y\big)\X^\top
    .
    \]
This is update~\eqref{eq:krasulina-kpca} without enforcing
the orthonormality constraint using the $\texttt{QR}$ decomposition and in which the projection is calculated using $\Cd$ instead of $\C^\top$.

A few additional remarks are in order. First, it may seem plausible to replace the projected gradient term $\pgrad\, \hat{\ell}_{\text{\tiny comp}}(\C)$ in~\eqref{eq:krasulina-kpca} with a more recent projected gradient $\pgrad\, \hat{\ell}_{\text{\tiny comp}}(\Ct)$ to achieve an implicit update
similar to~\eqref{eq:k-pca-implicit}. However, note that
$\Ct \notin \St$ in general and thus, this would not
correspond to a valid projected gradient update. Avoiding
the orthonormality constraint assures that the future
gradient is indeed a valid descent direction. Additionally,
note that~\eqref{eq:k-pca-implicit} is only partially implicit since we use the old $\C$ for finding the projection $\X$. Unfortunately, the fully implicit update does not yield a closed form solution.

The implicit update~\eqref{eq:krasulina-kpca} has a simple form in the stochastic setting,
when a single observation $\y_t$ is received at round $t$. Let $\x_t = \Cd\,\y_t$ denote the projection. Applying the Sherman-Morrison formula~\cite{golub} for the inverse of rank-one matrix updates,
we can write~\eqref{eq:k-pca-implicit} as
\begin{equation}
\boxed{
    \C^{\text{\tiny new}}
        = \C - \frac{\eta}{1 + \eta\,\Vert\x_t\Vert^2}\,\,
	(\C\,\x_t -\y_t)\, \x_t^\top}\label{eq:pca-single-m}\, .
\end{equation}
which we call the \emph{stochastic implicit Krasulina} update. Note that the learning rate $\frac{\eta}{1+\eta \Vert x_t\Vert^2}$
in~\eqref{eq:pca-single-m} is essentially inversely proportional to the norm of
the individual projection $\x_t$. The same fractional form of the learning
rate appears in the implicit update for online stochastic
gradient descent for linear regression~\cite{eg,pnorm}
which coincides with the differently motivated ``normalized LMS'' algorithm of \cite{hassibi}.
As noted in~\cite{pnorm}, implicit updates have slower
initial convergence (due to the smaller learning rate) and
smaller final error rate compared to the explicit
updates. Also, as a result of the adaptivity of the
learning rate, the implicit Krasulina update becomes less
sensitive to the initial choice for the learning rate. We
will show this in the experimental section.

  \begin{algorithm}[t!]
      \SetAlgoLined
    \SetKwInOut{Input}{input}
    \SetKwInOut{Output}{output}
    \caption{Stochastic Implicit Krasulina Algorithm}\label{alg:krasulina}
      \Input{\, data stream $\y_t\,, t = 1,\ldots, T$}
    \Output{\, projection matrix $\Proj = \C\Cd$}
    \BlankLine
      \textbf{initialize}\, $\C$\, and \textbf{set}\, $\Cd = (\C^\top\C)^{-1}\C^\top$
       \BlankLine
      \For{$t\leftarrow 1$ \KwTo $T$}{
          $\x_t \leftarrow \Cd\y_t$\\
         $\res_t \leftarrow \C\,\x_t-\y_t$\\
         $\eta_{\x_t} \leftarrow \sfrac{\eta_t}{(1 + \eta_t\,\Vert\x_t\Vert^2)}$\\
         $\C \leftarrow \C - \eta_{\x_t}\, \res_t\,\x_t^\top$\\
         $\Cd \leftarrow \text{\texttt{RankOnePinvUpdate}}(\C, \Cd, \eta_{\x_t}\res_t, \x_t)$
      }
      $\Proj \leftarrow \C\Cd$\\
      \Return $\Proj$
  \end{algorithm}

\section{Efficient Implementation}

We provide some insight on the efficient implementation of
the implicit Krasulina update.  While our update avoids
the costly \texttt{QR} step, it may impose extra computational
complexity when calculating the projection. Namely, the computational complexity of the stochastic update~\eqref{eq:pca-single-m} is dominated by the calculation of the matrix pseudo-inverse $\Cd = (\C^\top\C)^{-1}\C^\top$.  In order to implement the stochastic updates efficiently, we can exploit the fact that each iteration involves a rank-$1$ update on matrix $\C$. One approach to reduce the computational complexity would be to maintain the inverse matrix $\Sb \coloneqq (\C^\top\C)^{-1}$ and update it accordingly at every iteration. Note that a rank-$1$ update on $\C$ corresponds to a rank-$2$ update on $\Sb^{-1}$. Thus, the inverse can be carried out efficiently using the Woodbury matrix identity~\cite{woodbury}.
The one-time complexity of calculating the inverse is $\mathcal{O}(k^3)$, but can be significantly reduced by e.g. using a rectangular diagonal matrix as the initial solution. Thus, the overall complexity of the algorithm for the stochastic update~\eqref{eq:pca-single-m} becomes $\mathcal{O}(k\,d)$.

A computationally more efficient approach is to directly
keep track of the matrix $\Cd$ and apply the rank-1 update
for the Moore-Penrose inverse proposed in~\cite{meyer}. The
complexity of each rank-$1$ update amounts to
$\mathcal{O}(k\,d)$. Our implicit Krasulina update
is summarized in Algorithm~\ref{alg:krasulina}, where the operator \texttt{RankOnePinvUpdate} is described in~\cite{petersen}.

\section{Probabilistic $k$-PCA and Online EM}

We now provide an alternative motivation for our new
implicit Krasulina update~\eqref{eq:k-pca-implicit} as an
instantiation of the recent online Expectation Maximization
(EM) algorithm~\cite{online_em} for a certain probabilistic $k$-PCA model.
This probabilistic $k$-PCA model was introduced in~\cite{roweis,tipping} as a linear-Gaussian model:
\begin{equation}
\begin{aligned}
    \label{eq:prob-pca-model}
    \y = \C\, \x & + \nub\,,\,\,\,\\
    \text{ where }\,\,\, \x \sim \Nc(\zero_k,\, \I_k) \,\,\,& \text{ and }\,\,\,\nub \sim \Nc(\zero_d,\,\Q)\, .
    \end{aligned}
\end{equation}
where $\x$ is the $k \times 1$ unknown hidden state, $\nub$ denotes the $d\times 1$ observation noise, and $\C \in \RR^{d\times k}$.  Also, $\Nc(\mub,\mathrm{S})$ denotes a Gaussian distribution having mean $\mub$ and covariance $\mathrm{S}$. Usually, the noise covariance is assumed to be isotropic, i.e. $\Q = \epsilon\,\I_d$. Note that since all random variables are Gaussian, the posterior distribution of the hidden state also becomes Gaussian, that is,
\begin{align*}
    \x|\,\y \sim \Nc(\bet\, \y,\, \I_k - \bet\, \C)\, ,
    \text{ where }\bet = \C^\top(\C\,\C^\top + \Q)^{-1}.
\end{align*}
 An interesting case happens in the limit where the covariance of the noise $\nub$ becomes infinitesimally small. Namely, in the limit $\Q = \lim_{\epsilon \rightarrow 0} \epsilon\,\I_d$, the likelihood of a point $\y$ is solely determined by the squared Euclidean distance between $\y$ and its reconstruction $\C\,\x$.  The posterior of the hidden state collapses into a single point,
 \[
     \x|\,\y \sim \Nc\big((\C^\top\,\C)^{-1}\C^\top \y,\, \zero\big)\, = \delta\big(\x - (\C^\top\,\C)^{-1}\C^\top \y\big)\, ,
\]
 where $\delta$ is the Dirac measure. Moreover, the maximum
 likelihood estimator for $\C$ is achieved when $\C$ spans the space of top-$k$ eigenvectors of the data covariance matrix~\cite{tipping}.
 Thus, the linear-Gaussian model reduces to the vanilla $k$-PCA problem.

 Using the probabilistic $k$-PCA formulation allows solving the vanilla $k$-PCA problem iteratively in the zero noise limit via the application of the EM algorithm~\cite{dempster}. Let $\Th = \{\epsilon, \C\}$
denote the set of parameters of the probabilistic latent
variable model with joint probability density
$P_{\Th}(\x,\y)$. The EM upper-bound can be written as
 \begin{align}
    &\UP_{\Th^{(t)}}(\Tht|\Y^{(t)})\! = -\sfrac{1}{N^{(t)}}\sum_n \!\int_\x\! P_{\Th^{(t)}}(\x|\,\y_n)\log P_{\Tht}(\x,\y_n)\nonumber\\
     &= \frac{1}{2}\epstinv \tr\big(\Ct\Sig\Ct^\top - \frac{2}{N}\Y\X^\top\Ct^\top\big)
    + d\log\,\epst + \text{const.}\label{eq:up}
    \end{align}
where $\X = \bet\,\Y$\, with\, $\bet = \C^\top(\C\,\C^\top + \epsilon\,\I_k)^{-1}$ and $\Sig = \I_k - \bet\C + \frac{1}{N}\,\X\X^\top$. We now consider the case when $\epsilon$ becomes infinitesimally small. Note that in the limit $\epsilon \rightarrow 0$, we have $\bet = (\C\,\C^\top)^{-1}\,\C^\top = \Cd$ and $\bet\,\C = \I_k$.
 Iteratively forming the EM upper-bound and minimizing it yields the following procedure\footnote{Assuming that $\X$ is rank-$k$.}~\cite{roweis}:
 \begin{equation}
 \begin{aligned}
     \label{eq:pca-batch-em}
     \X & = \Cd\Y=(\C\,\C^\top)^{-1}\,\C^\top\,\Y\,\,\,&& \text{ (E-step)\, ,}\\
     \C^{\text{new}} & = \Y\X^\dagger=\Y\X^\top(\X\X^\top)^{-1}\,\,\,&& \text{ (M-step)\,,}
 \end{aligned}
 \end{equation}
where the matrices $\X$ and $\Y$ are defined as before. Because of the tightness of the EM upper-bound, every step of the EM algorithm is guaranteed to either improve the compression loss or leave it unchanged~\cite{online_em}. The final projection matrix is obtained as $\Proj = \C\Cd$.

We now motivate our new implicit Krasulina update as online version of the update~\eqref{eq:pca-batch-em}. This can be achieved by an application
of a recent online EM algorithm developed in~\cite{online_em}.
In the online setting, the learner receives a mini-batch of
observations (usually a single observation) at a time and
performs parameter updates by minimizing the negative
log-likelihood of the given examples. In order to make the
learning stable, an inertia (aka regularizer) term is added
to the loss to keep the updates close to the old
parameters. Thus, the learner minimizes the combined
inertia plus loss of the current iteration. Let $\Y^{(t)}\in \RR^{d\times
N^{(t)}}$ be the given batch of observations at round $t$.
The online EM algorithm introduced in~\cite{online_em} is
motivated in the same manner by minimizing the following
loss at iteration $t$:
\[
    \Th^{(t+1)}  = \argmin_{\Tht}\, \Big(\sfrac{1}{\eta^{(t)}}\, \Delta_{\text{RE}}(\Th^{(t)}, \Tht) +  \UP_{\Th^{(t)}}(\Tht|\,\Y^{(t)})\Big).
    \]
    where
    $\Delta_{\text{RE}}(\Th, \Tht) = \int_{\x,\,\y}
    P_{\Th}(\x,\y)\log\frac{P_{\Th}(\x,\y)}{P_{\Tht}(\x,\y)}$
    is the relative entropy divergence between the joints and
     $\eta^{(t)} > 0$ is the learning rate. In the following, we consider one iteration of the online EM algorithm and drop the $t$ superscript to avoid clutter\footnote{We also use $\etap$ instead of $\eta$ for reasons which will be clear in the following.}.

We now consider the online EM algorithm when applied to the probabilistic $k$-PCA model~\eqref{eq:prob-pca-model}. Note that $\Th = \{\epsilon, \C\}$. We can write
\begin{equation}
\begin{aligned}
    \Delta_{\text{RE}}(\Th, \Tht) & = \frac{d}{2} \big(\frac{\epsilon}{\epst} - \log \frac{\epsilon}{\epst} - 1 \big)\\ & + \frac{1}{2}\,\epstinv\, \tr\big((\C - \Ct)\,(\C - \Ct)^\top\big)\, ,\label{eq:inerta}
    \end{aligned}
    \end{equation}
We now consider the case where $\epst = \epsilon$ is fixed. Combining~\eqref{eq:inerta} and~\eqref{eq:up}, the update for parameter $\C$ becomes
\begin{multline*}
    \C^{\text{\tiny new}} = \Big(\frac{1}{N} \Y\, \X^\top + \frac{\epsilon^{-1}}{\etap}\, \C\Big)\\
    \times\, \Big(\I_k - \bet\,\C + \frac{1}{N}\,\X\,\X^\top +  \frac{\epsilon^{-1}}{\etap}\, \I_k\Big)^{-1} \,\,\,\,\text{(M-step)}\, .
\end{multline*}
In order to recover the updates for the online $k$-PCA, we again need to consider the case when $\epsilon$ becomes infinitesimally small. Choosing $\etap(\epsilon)$ such that $\lim_{\epsilon \rightarrow 0}\sfrac{\epsilon^{-1}}{\etap} = \sfrac{1}{\eta}$\, yields~\eqref{eq:k-pca-implicit}. Note that the E-step of the online EM algorithm becomes identical to~\eqref{eq:pca-batch-em}. Also, the limit case $\eta \rightarrow 0$ keeps the parameters unchanged, i.e. $\C^{\text{\tiny new}} = \C$ and the case $\eta \rightarrow \infty$ recovers the batch EM updates~\eqref{eq:pca-batch-em}.
\section{Distributed Setting}

The alternative view of the implicit Krasulina update~\eqref{eq:k-pca-implicit} as an EM step allows combining multiple models in a distributed setting via the online EM framework introduced in~\cite{online_em}. More specifically, given a set of $M$ hidden variable models parameterized by $\{\Th^{(i)}\}_{i=1}^M$, the optimal combined model $\Th^*$ corresponds to the minimizer of the following objective
\[
\Th^* = \argmin_{\Th}\, \sum_{i \in [M]} \alpha_i\,\Delta_{\text{RE}}(\Th^{(i)}, \Th)\, ,
\]
where $\alpha_i \geq 0$ is the weight associated with model $i$. The weight of the model can be assigned based on the performance on a validation set or based on the number of observations processed by the model so far. For synchronous updates, we simply have $\alpha_i = \frac{1}{M}$. In a distributed online PCA setting, let $\C^{(i)}$ denote the matrix learned by the model $i$. By fixing $\epsilon$ for all the models and letting $\Th^{(i)} = \{\epsilon, \C^{(i)}\}$, we have
\begin{equation}
    \label{eq:combine}
    \C^* = \frac{\sum_{i \in [M]} \alpha_i\, \C^{(i)}}{\sum_{i \in [M]} \alpha_i}\, .
\end{equation}
The probabilistic view of our implicit Krasulina updates
allows training multiple $k$-PCA models in parallel and
combining them efficiently via simple averaging.
Note that for the previous approaches, the $\C^{(i)},\, i\in [M]$
matrices are orthonormal and to the best of our knowledge, there is no systematic way of combining rank-$k$ orthonormal matrices. One trick would be to also average these matrices.  However, the average of a set of orthonormal matrices does not necessarily yields an orthonormal matrix and a \texttt{QR} step is required after combining.
As we will show experimentally, the heuristic of simply
averaging the orthonormal matrices $\C^{(i)}$ produced by the Krasulina
and Oja updates
and then orthonormalizing the average yields poor empirical results.
On the other hand, our new implicit Krasulina update
produces arbitrary matrices $\C^{(i)}\in \R^{d\times k}$
and averaging those matrices (as advised by the online EM
framework) results in excellent performance.

\section{Experiments}
In this section, we perform experiments on real-world datasets using our proposed implicit form of Krasulina's update~\eqref{eq:pca-single-m} and contrast the results with Oja's, Krasulina's, and the incremental algorithm~\cite{incremental}. We perform experiments on  MNIST dataset\footnote{\url{http://yann.lecun.com/exdb/mnist/}} of \num{70000} handwritten images with dimension \num{784} and CIFAR-10 dataset\footnote{\url{https://www.cs.toronto.edu/~kriz/cifar.html}} of real-world images of dimension \num{3072} having \num{60000} images in total. We apply all the updates in a stochastic manner by sweeping over the data once. We consider a decaying learning rate of $\eta = \sfrac{\eta_0}{t^\gamma}$\, where $t$ denotes the iteration number, $0.5 \leq \gamma < 1$ is a constant, and $\eta_0$ denotes the initial learning rate. For each dataset, we randomly select 10\% of the data as a validation set to select the optimal initial learning rate and the value of $\gamma$ for each algorithm.  We use $\gamma = 0.9$ for Krasulina's and Oja's methods and set $\gamma = 0.8$ for our method. Note that the incremental algorithm does not require a learning rate. We report the performance on the full dataset. We repeat each experiment 10 times with a different random initialization and report the average. For comparison, we also calculate the result of the batch PCA solution using SVD decomposition of the empirical data covariance matrix for each experiment.

\begin{figure*}[t!]
\vspace{-0.5cm}
\begin{center}
\subfigure{\includegraphics[width=0.24\textwidth]{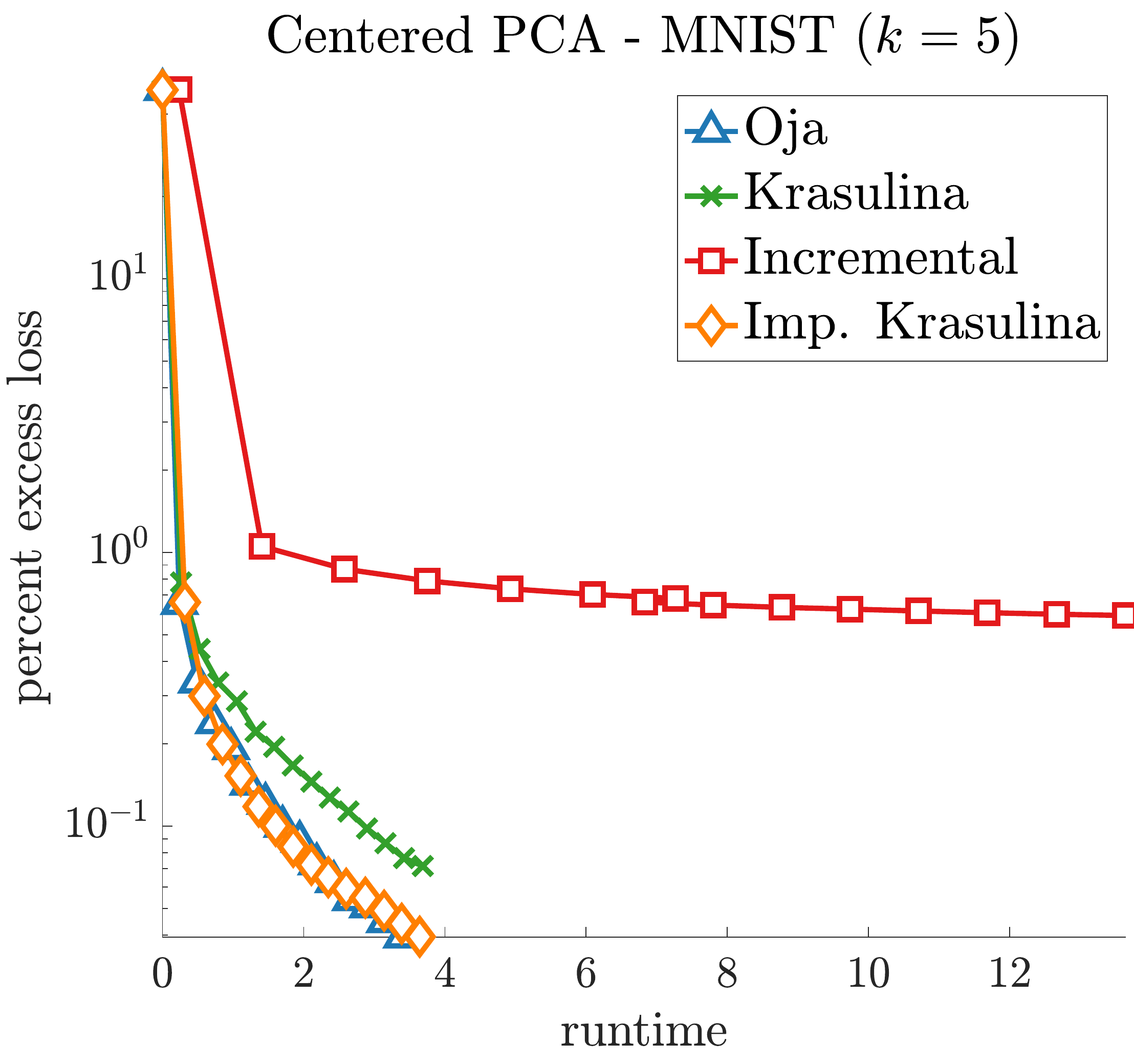}}\,
    \subfigure{\includegraphics[width=0.24\textwidth]{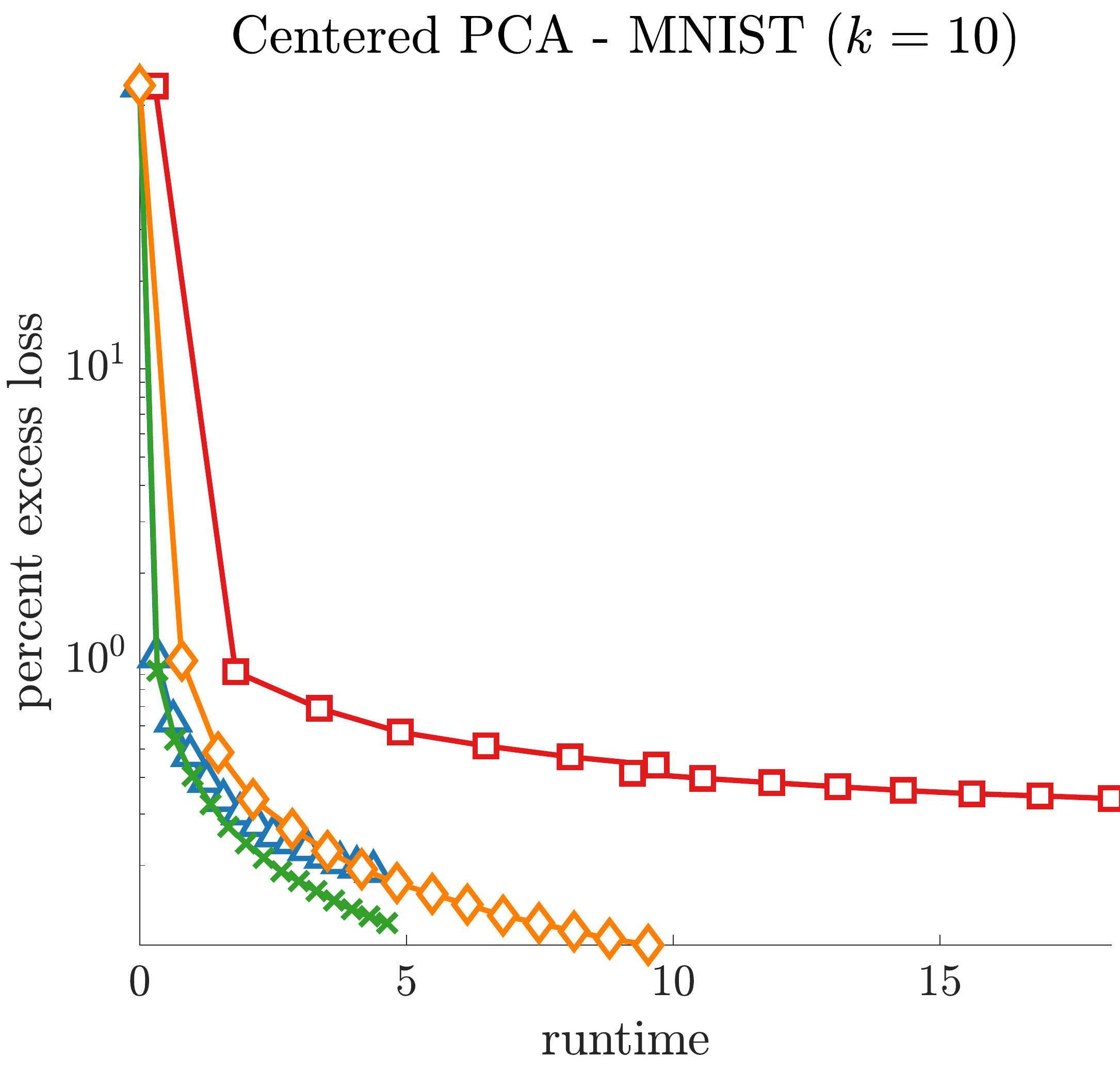}}\,
    \subfigure{\includegraphics[width=0.24\textwidth]{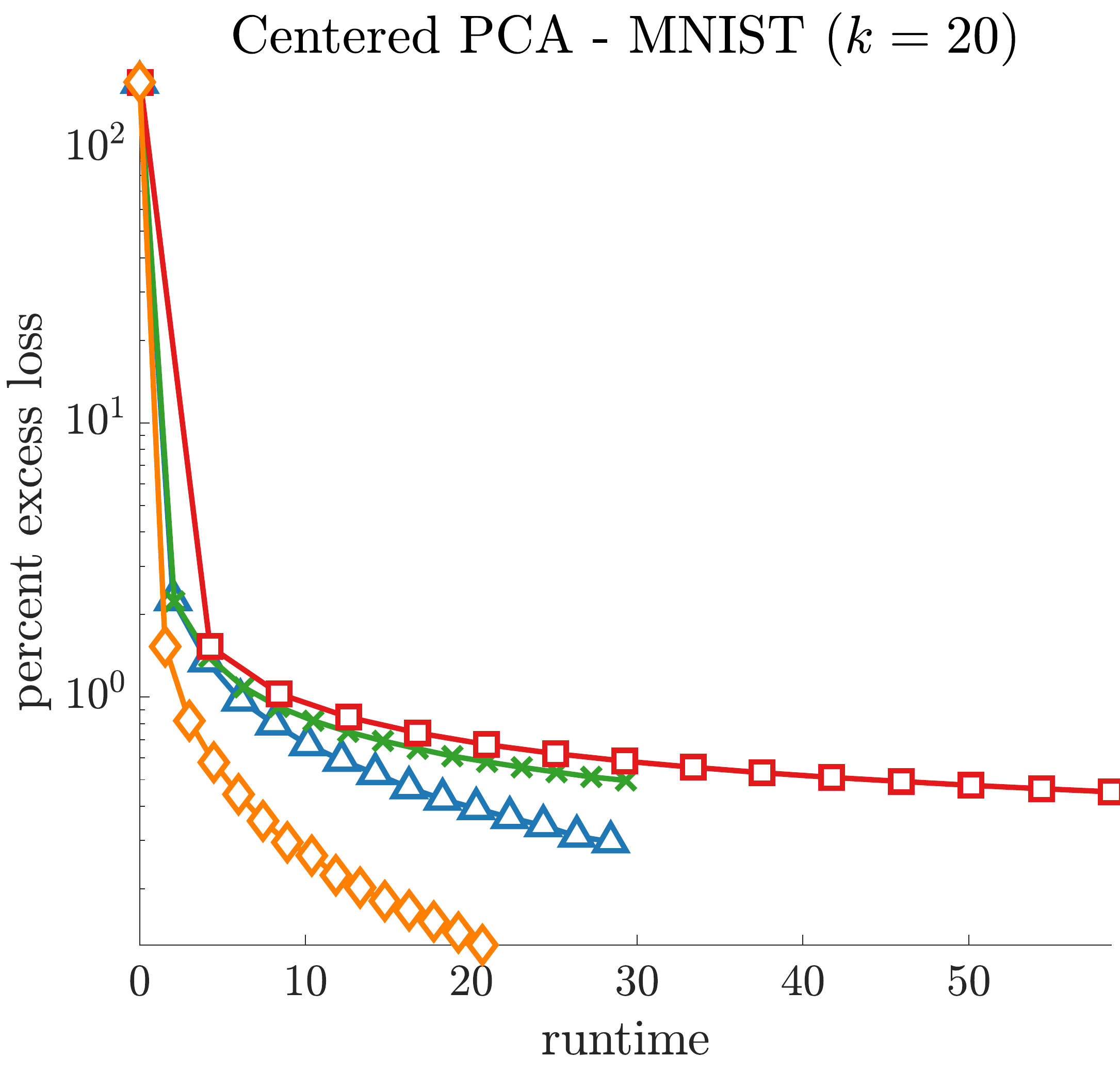}}\,
    \subfigure{\includegraphics[width=0.24\textwidth]{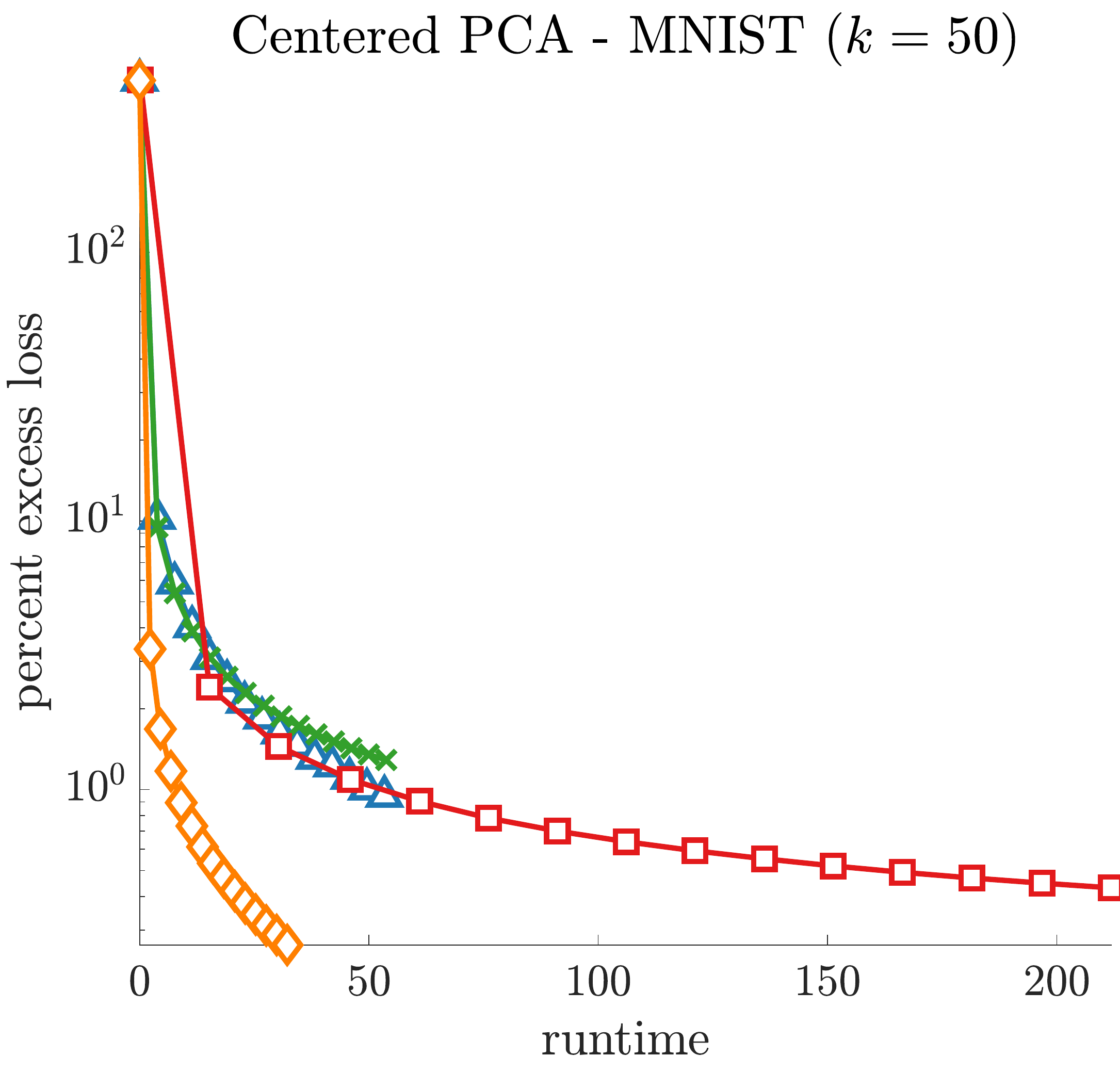}}\,\hfill\\
    \subfigure{\includegraphics[width=0.24\textwidth]{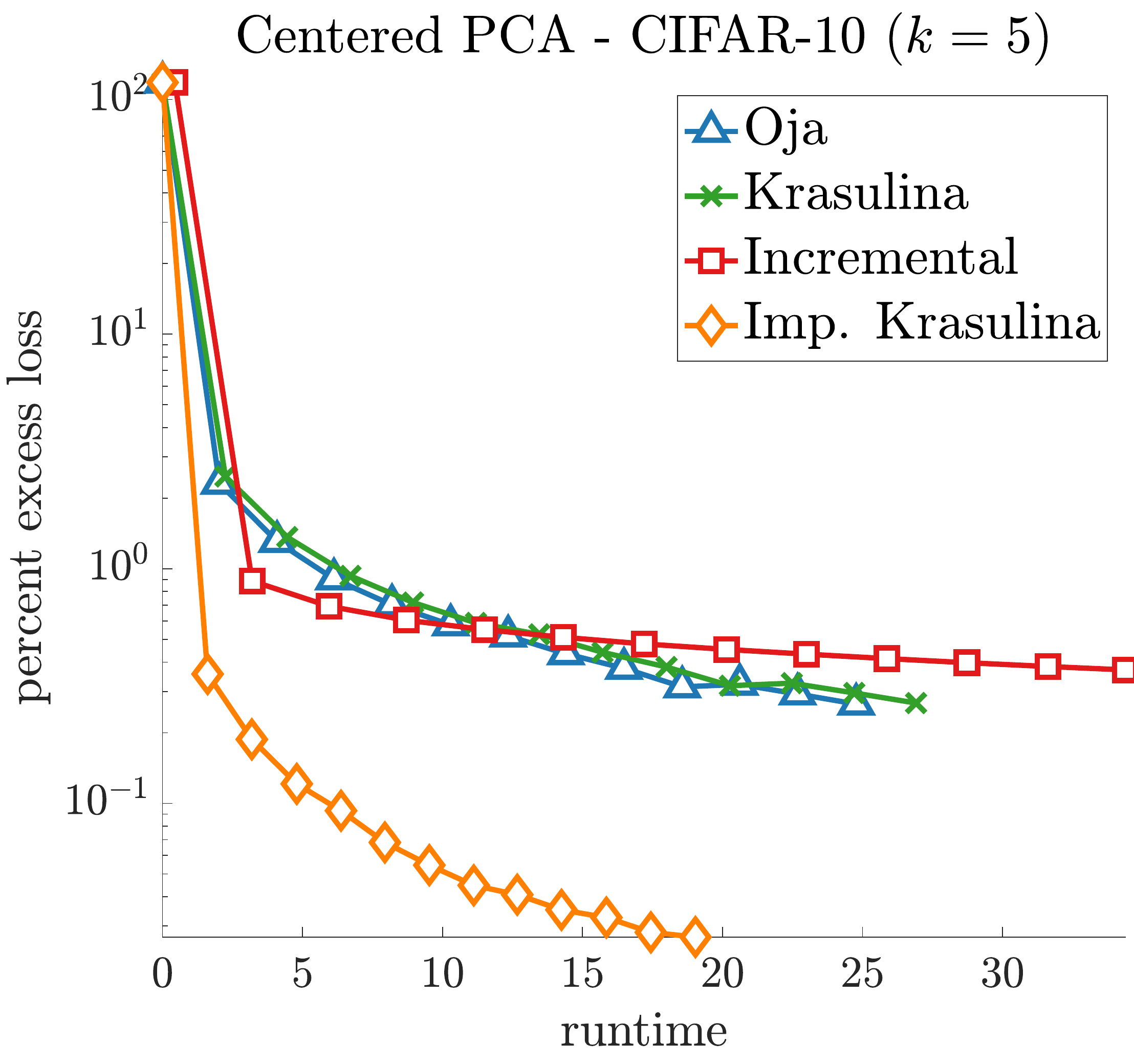}}\,
    \subfigure{\includegraphics[width=0.24\textwidth]{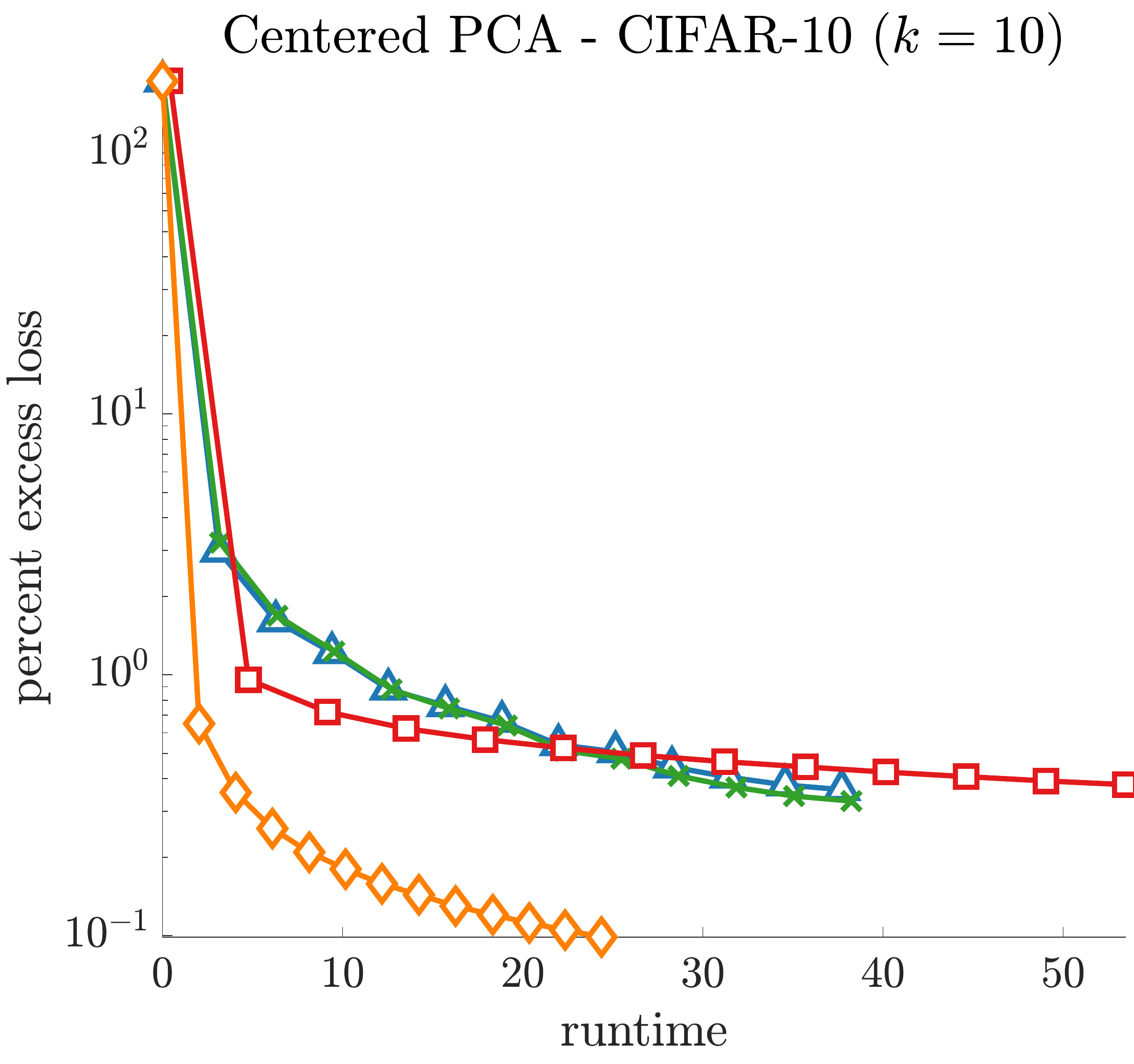}}\,
    \subfigure{\includegraphics[width=0.24\textwidth]{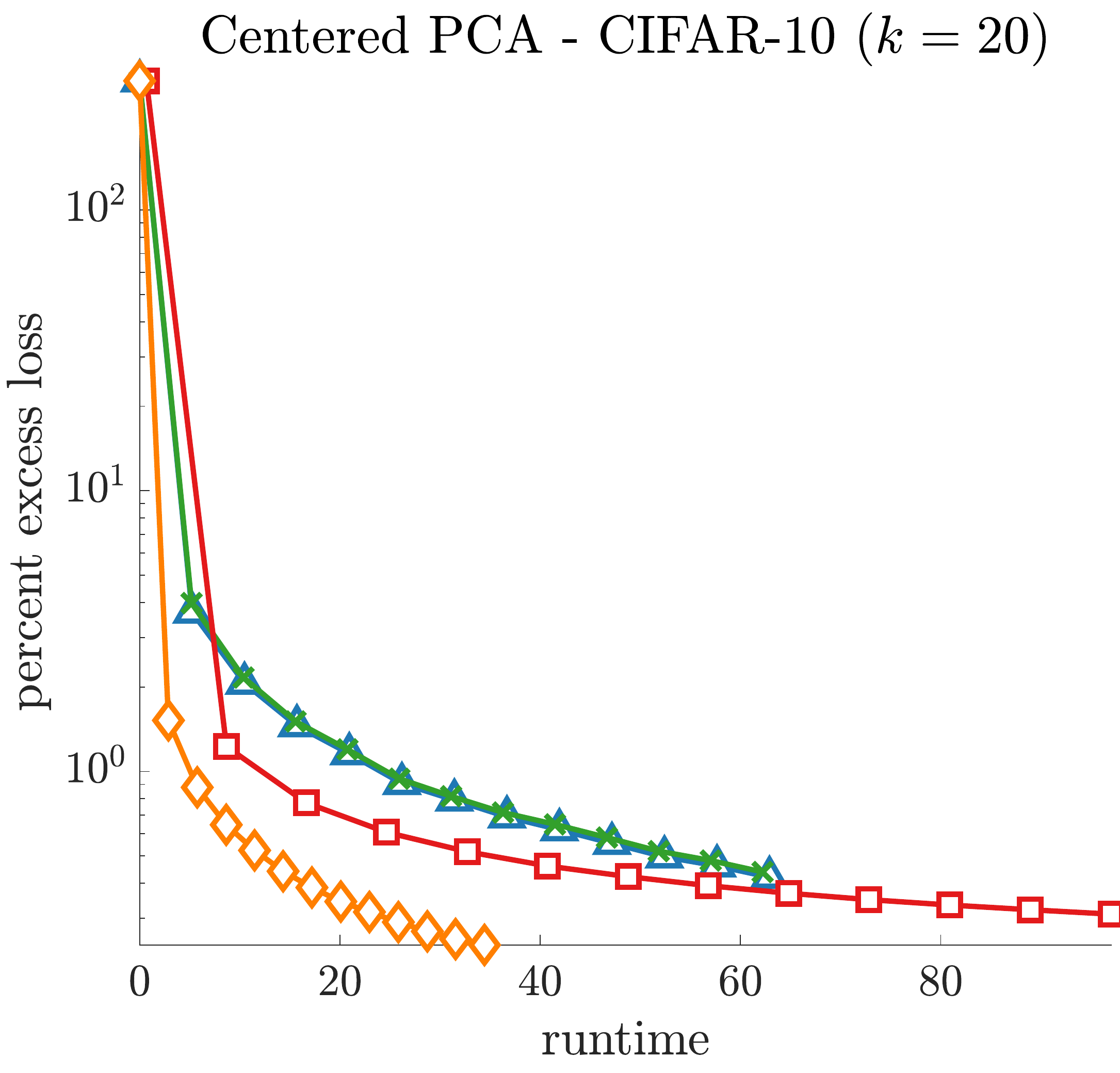}}\,
    \subfigure{\includegraphics[width=0.24\textwidth]{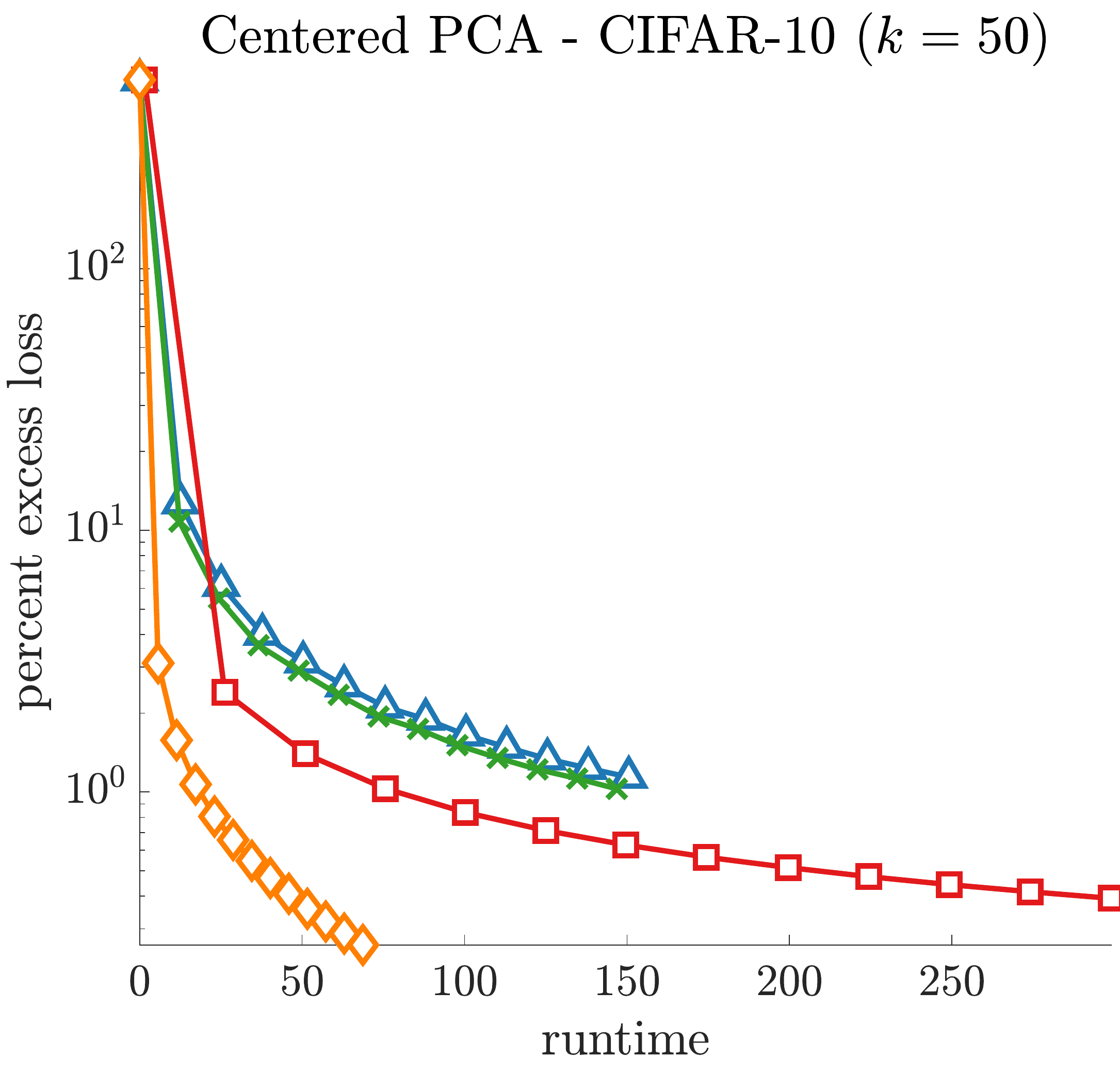}}\,\hfill
     \caption{Centered online $k$-PCA: the results of different algorithms on the MNIST (top) and CIFAR-10 (bottom). The value of $k$ is shown on top of each figure. We plot the percentage of the excess loss which is defined as the normalized regret w.r.t to the best offline comparator (i.e. full-batch PCA). Each dot shows the progress of the algorithm in 1000 iteration intervals. Our proposed implicit Krasulina algorithm achieves the best convergence and provides the best runtime overall, especially when the values of $d$ and $k$ are large.}\label{fig:centered}
\end{center}
\end{figure*}
\begin{table*}[t!]
\vspace{-0.1cm}
{\small
\begin{center}
\resizebox{0.96\textwidth}{!}{
\begin{tabular}{lccccccc}
\toprule
\multirow{2}{*}{Method} & \multirow{2}{*}{$\eta_0$-Scale} & \multicolumn{3}{c}{MNIST} & \multicolumn{3}{c}{CIFAR10}\\
\cmidrule(lr){3-5}\cmidrule(lr){6-8}
& & $k=5$ & $k=10$ & $k=20$ & $k=5$ & $k=10$ & $k=20$\\
\midrule
\midrule
Batch PCA & -- & $35.16$ & $26.95$ & $18.74$ & $86.98$ & $65.70$ & $48.65$\\
\midrule
\multirow{3}{*}{Oja} & \multirow{1}{*}{$0.1\times$} & $35.79\pm 0.38$ & $29.78\pm 0.38$ & $21.55\pm 0.25$ & $116.26\pm 2.58$ & $66.81\pm 0.58$ & $58.28 \pm 0.65$\\
 &\multirow{1}{*}{$1\times$} & $35.19 \pm 0.05$ & $26.98 \pm 0.02$ & $18.80\pm 0.03$ & $87.28\pm 0.43$ & $65.90\pm 0.01$ & $48.86 \pm 0.05$\\
            & \multirow{1}{*}{$10\times$} & $35.29\pm 0.01$ & $27.08 \pm 0.01$ &  $19.01\pm 0.01$ & $87.22\pm 0.04$ & $67.68\pm 0.13$ & $50.36 \pm 0.13$\\
\midrule
\multirow{3}{*}{Krasulina} & \multirow{1}{*}{$0.1\times$} & $35.78\pm 0.37$ & $29.92\pm 0.37$ & $21.44\pm 0.24$ & $116.86\pm 2.58$ & $66.76\pm 0.65$ & $57.62\pm 0.65$\\

            &\multirow{1}{*}{$1\times$} & $35.17\pm 0.00$ & $26.96 \pm 0.01$ & $18.79\pm 0.02$ & $87.04\pm 0.52$ & $65.90\pm 0.01$ & $48.87 \pm 0.05$\\
            & \multirow{1}{*}{$10\times$} & $35.30\pm 0.01$ & $27.09\pm 0.01$ & $19.02\pm 0.02$ & $87.17\pm 0.04$ & $67.87\pm 0.16$ & $50.56\pm 0.15$\\
\midrule
Incremental & -- & $35.26\pm 0.10$ & $27.01 \pm 0.04$ & $18.83\pm 0.05$ & $87.50\pm 0.46$ & $65.75\pm 0.05$ & $48.82\pm 0.10$\\
\midrule
\multirow{3}{*}{Imp. Krasulina} & \multirow{1}{*}{$0.1\times$} & $35.17\pm 0.01$ & $26.96\pm 0.01$ & $18.78\pm 0.02$ & $87.00\pm 0.00$ & $65.75\pm 0.05$ & $48.75\pm 0.03$\\
 &\multirow{1}{*}{$1\times$} & $35.17\pm0.01$ & $26.97\pm 0.03$ & $18.77\pm 0.02$ & $87.01\pm 0.00$ & $65.78\pm 0.09$ & $48.74 \pm 0.04$\\
            & \multirow{1}{*}{$10\times$} & $35.17\pm0.01$ & $26.98 \pm 0.05$ & $18.77\pm 0.01$ & $87.02\pm 0.04$ & $65.74 \pm 0.01$ & $48.76\pm 0.03$\\

\bottomrule
\end{tabular}
}
\end{center}
}
\caption{Compression loss on different datasets using the optimal initial learning rate $\eta_0$ (selected using a validation set) and its scaling. Note that due to the adaptive form of learning rate, the results of our implicit Krasulina update is more stable w.r.t. to the choice of the initial learning rate.}
\label{tab:sensistivity}
\end{table*}
\subsection{Centered PCA}
We show the results of the algorithms on the centered datasets in Figure~\ref{fig:centered} as a function of runtime. We plot the percentage of excess loss (i.e. normalized regret) w.r.t to the best achievable loss by an offline algorithm (i.e. full-batch PCA). Each dot shows the progress of the algorithm in 1000 iteration intervals. As can be seen, our implicit Krasulina update achieves the best convergence among all the algorithms. Additionally, our algorithm is considerably faster in most cases. Especially, the advantage of our updates becomes more evident as the values of $d$ and $k$ increase. This can be explained by the low complexity of matrix updates for our algorithm versus the costly \texttt{QR} update or the eigen-decomposition operation for the remaining algorithms.

\begin{figure*}[t!]
\vspace{-0.5cm}
\begin{center}
    \subfigure{\includegraphics[width=0.24\textwidth]{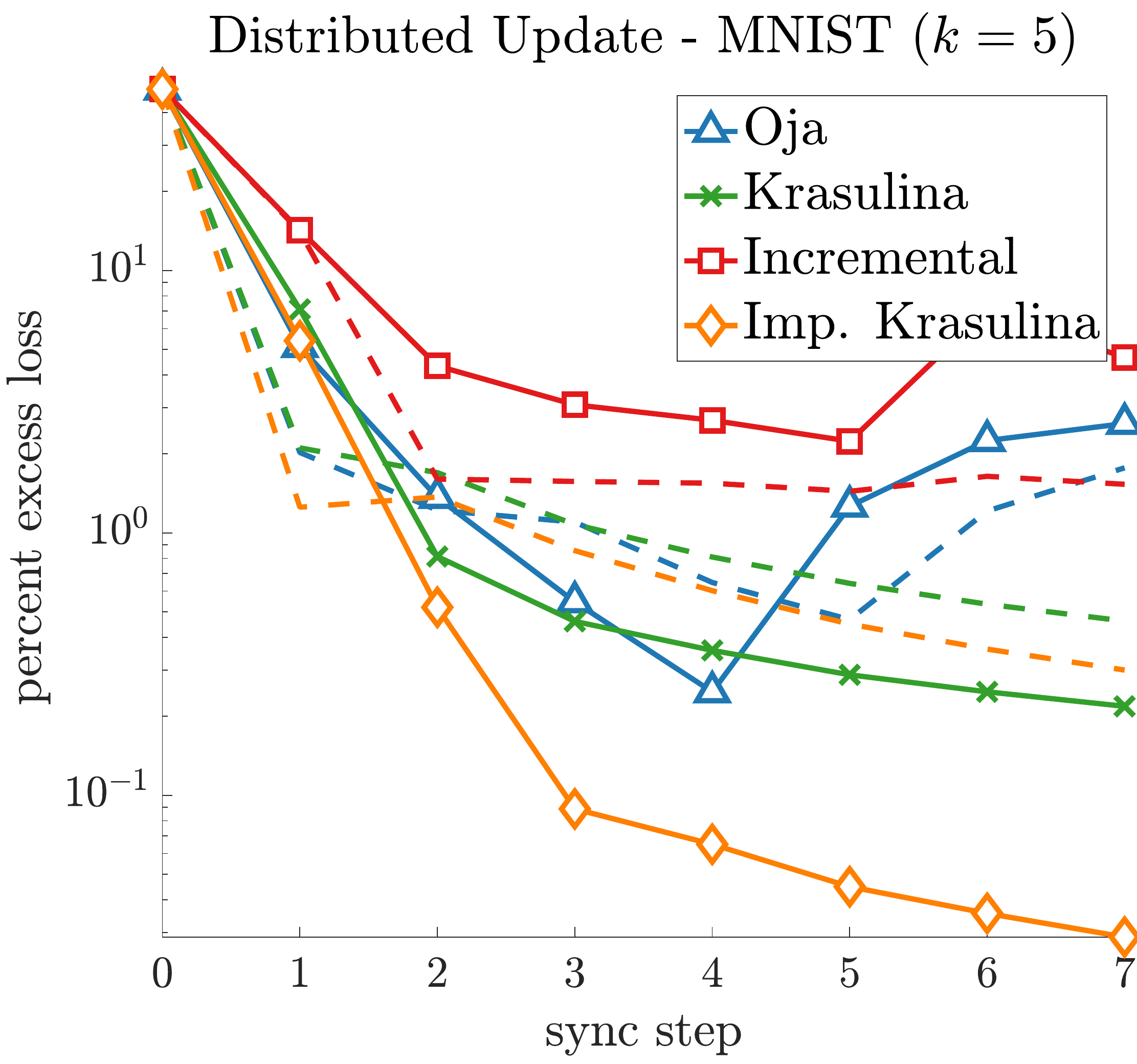}}\,
    \subfigure{\includegraphics[width=0.24\textwidth]{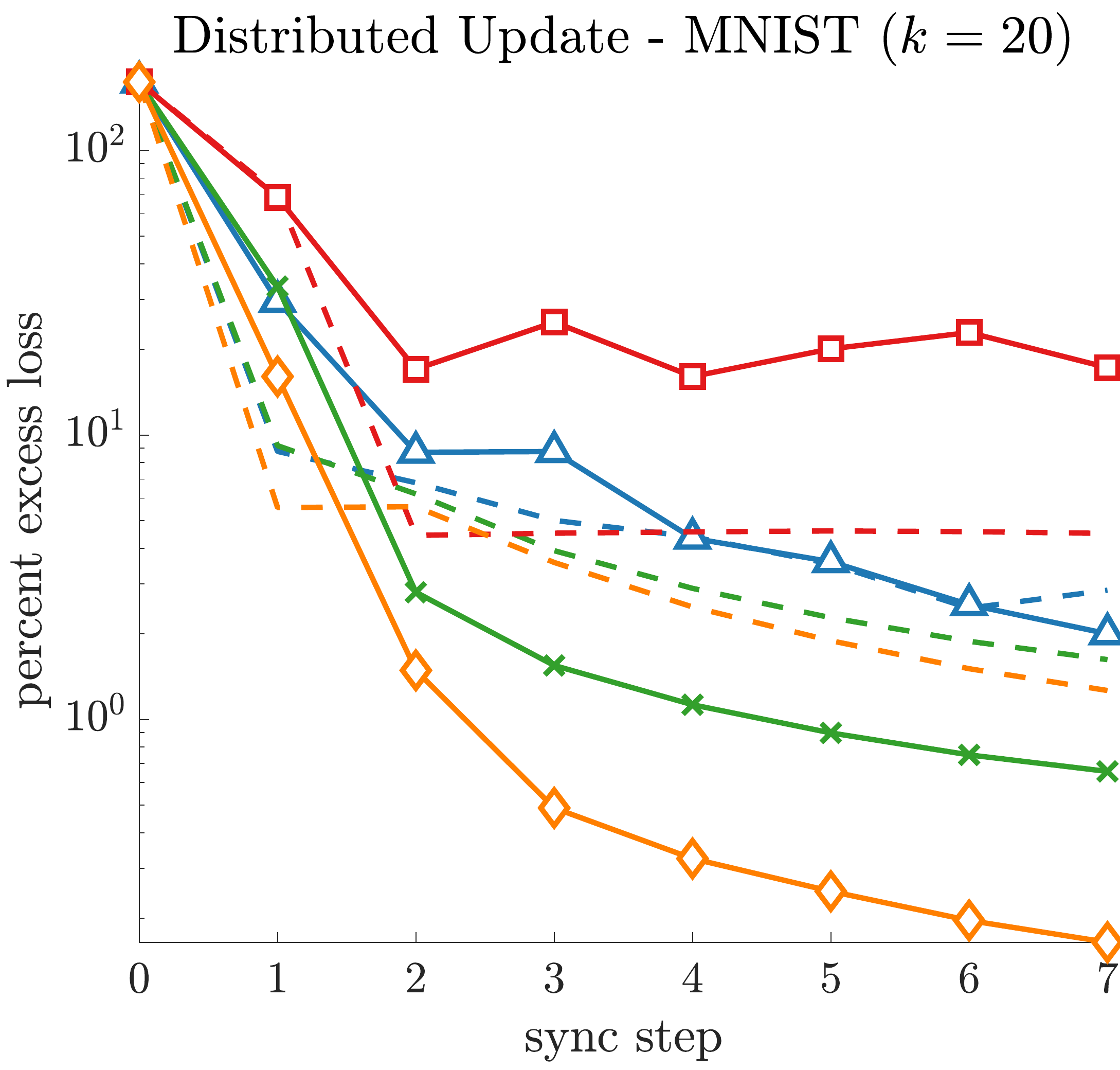}}\,
    \subfigure{\includegraphics[width=0.24\textwidth]{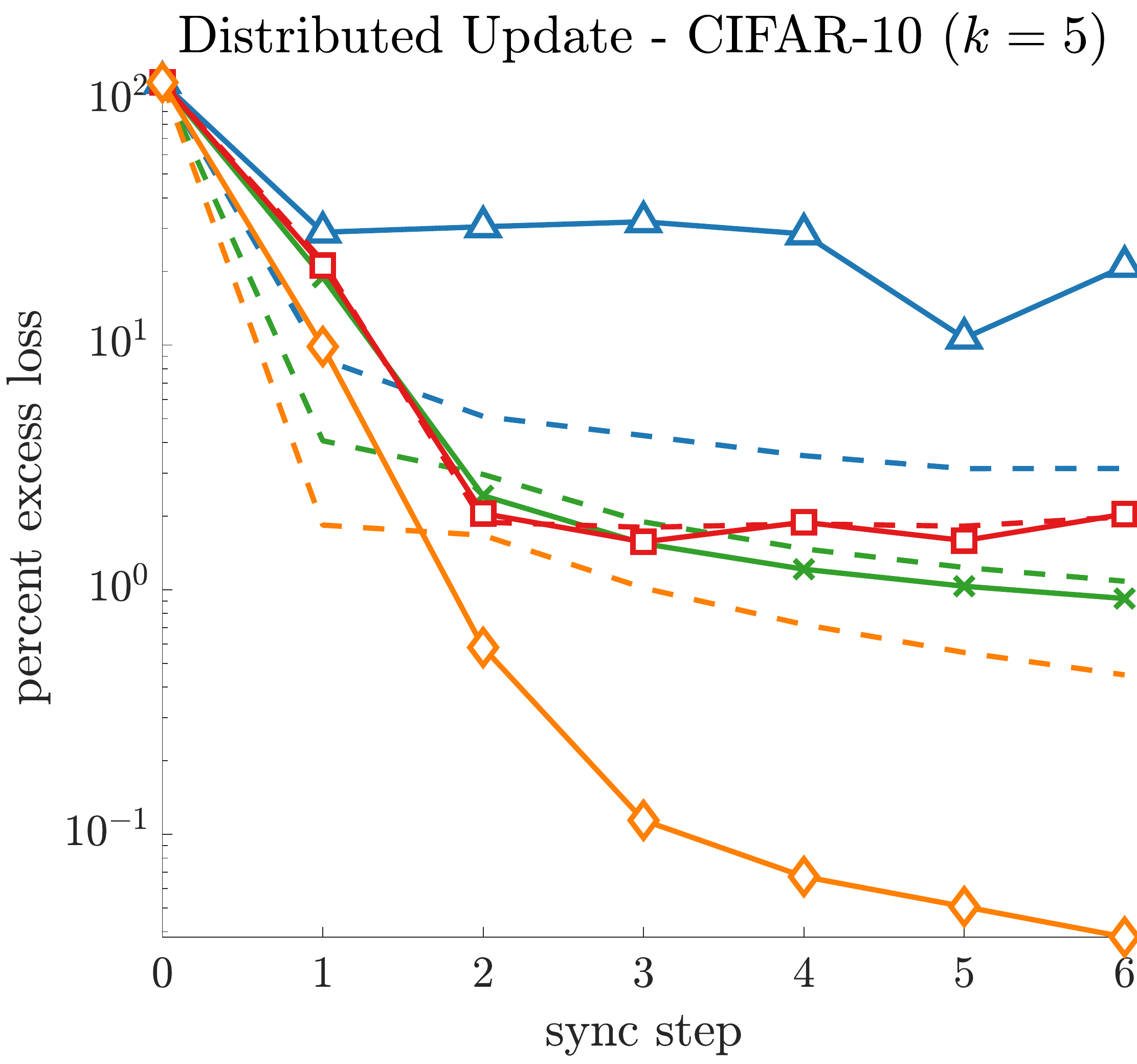}}\,
    \subfigure{\includegraphics[width=0.24\textwidth]{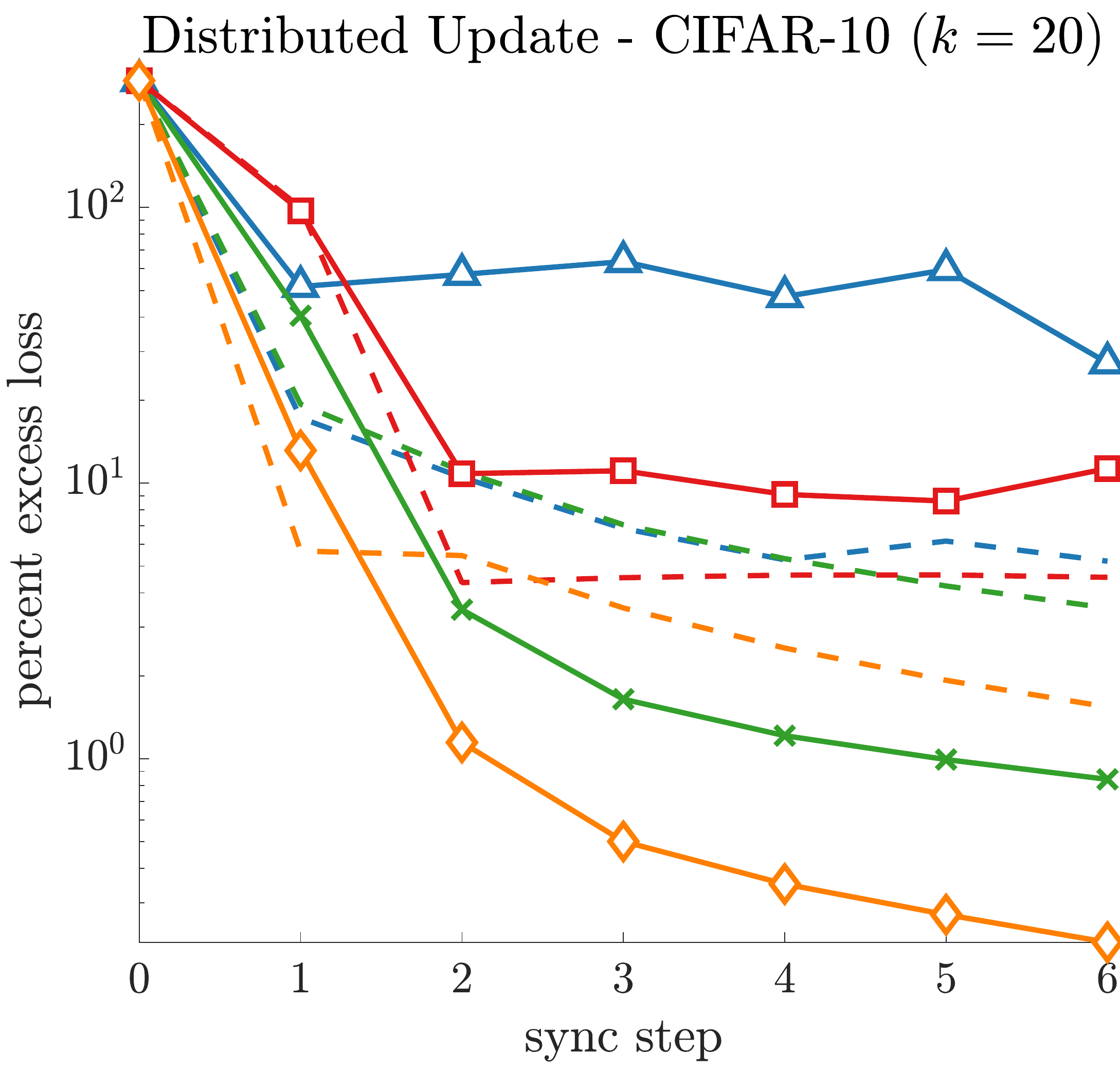}}\,\hfill
     \caption{Distributed setting: the results of different algorithms in a distributed setting. The value of $k$ is shown on top of each figure. The updates are carried out across $M = 10$ machines and the models are combined and propagated back every $1000$ iterations. The average loss of the machines for each algorithm is shown with a dashed line. The loss of the combined model is shown with a solid line. Note that our implicit Krasulina update consistently improves over time by combining the partial model. The remamining methods are less stable and provide poor results.}\label{fig:distributed}
\end{center}
 \vspace{-0.2cm}
\end{figure*}
\begin{table}[t!]
\setlength{\tabcolsep}{4pt}
\vspace{-0.1cm}
\begin{center}
\resizebox{0.48\textwidth}{!}{
\begin{tabular}{lcccc}
\toprule
\multirow{2}{*}{Method} &  \multicolumn{2}{c}{MNIST} & \multicolumn{2}{c}{CIFAR10}\\
\cmidrule(lr){2-3}\cmidrule(lr){4-5}
 & $k=5$ & $k=20$ & $k=5$ & $k=20$\\
\midrule
\midrule
Batch PCA & $35.16$ & $18.74$ & $86.98$ & $48.65$\\
\midrule
Single & $37.17\pm0.01$ & $18.77\pm 0.02$ & $87.01\pm 0.00$ & $48.74\pm0.04$ \\
\midrule
Parallel & $35.17\pm0.00$ & $18.77\pm0.01$ & $87.02\pm0.00$ & $48.76\pm0.04$ \\
\bottomrule
\end{tabular}
}
\end{center}

\caption{Loss of the optimal algorithm (i.e. batch PCA) and the loss of the implicit Krasulina algorithm obtained by running on a single machine vs. running in parallel ($M=10$).}
\label{tab:distributed}
\vspace{-0.2cm}
\end{table}

\subsection{Sensitivity to Initial Learning Rate}
We demonstrate the sensitivity of each algorithm to the choice of initial learning rate by applying each algorithm using the optimal learning rate (obtained based on the performance on a validation set) as well as the results of running the same algorithm with $0.1\times$ and $10\times$ the optimal learning rate value. We show the results in Table~\ref{tab:sensistivity}. In the table, we show the final loss of each algorithm on the full dataset. Note that the incremental algorithm is unaffected since no learning rate is used for this method. Among the other methods, our implicit Krasulina algorithm provides excellent convergence even with the non-optimal learning rate and has the lowest sensitivity. The performance of Oja's and Krasulina's algorithms immediately deteriorates as the value of the learning rate is altered.

\subsection{Distributed Setting}
Finally, we evaluate the results of different algorithms in a distributed setting. We randomly split the data across $M = 10$ machines and perform synchronous updates by combining the results every 1000 iterations. In our experiments all $M$ sub-problems have the same size and
we use $\alpha_i=\sfrac{1}{M}$. We propagate back the value of the combined matrix to each machine. For our proposed algorithm, we apply the online EM framework~\cite{online_em} which corresponds to averaging the learned matrices of all machines as in~\eqref{eq:combine}. For all the remaining methods, since there exists no clear procedure for combining orthonormal matrices, we na\"ively apply the same procedure. However, the average of a set of orthonormal matrices does not necessarily correspond to an orthonormal matrix. Thus, we apply the $\texttt{QR}$ step on the combined matrix before calculating the loss and propagating back the combined value. We show the results in Figure~\ref{fig:distributed}. In the figure, we plot the percentage of the excess loss w.r.t. to the loss of the optimal algorithm, i.e. batch PCA. The solid line indicates the performance of the combined model. We also calculate the loss of each individual model (of each machine) on the full dataset over time and plot the average loss value across the machines with a dashed line. This verifies that whether the combined model performs better than each individual model on average.

As can be seen from the figure, our implicit Krasulina algorithm consistently provides excellent performance and converges to the optimal solution. Among the remaining methods, Krasulina's algorithm provides better convergence behaviour, but converges to an inferior solution. The Oja algorithm as well as the incremental algorithm fail to provide comparable results. The final loss of the combined model for our algorithm is also very close to the final loss of running our algorithm on a single machine on the full dataset, as shown in Table~\ref{tab:distributed}.


\section{Conclusion}

The advantage of using future gradients has now appeared
in a large variety of contexts (e.g.~\cite{nesterov,vishy,bartlett}).
Here we develop a partially implicit version of Krasulina's
update for the centered $k$-PCA problem that is dramatically
better than the standard explicit update.
A second key component is to use a latent variable
interpretation of the problem and then apply the online EM
framework for combining models in a distributed setting
\cite{online_em}.
Combining sub-models is equally important for the $k$-PCA
problem, and finding further practical applications
of this second component and blending it with other methods
is a promising future direction.


\bibliography{refs}
\bibliographystyle{aaai}

\end{document}